%% file: main.tex
\documentclass[letterpaper,10pt,conference]{ieeeconf}
\IEEEoverridecommandlockouts
\usepackage{cite}
\usepackage{amsmath,amssymb,amsfonts}
\usepackage{algorithmic}
\usepackage{graphicx}
\usepackage{textcomp}
\usepackage{xcolor}
\usepackage{algorithm}
\usepackage{multirow}
\let\labelindent\relax
\usepackage{enumitem}
\usepackage[hyphens]{url}
\usepackage[hidelinks]{hyperref}
\usepackage{subcaption}
\usepackage{textcomp}

\usepackage{color}

\usepackage{wrapfig}
\usepackage{soul}
\usepackage{threeparttable}
\usepackage{setspace}
\usepackage{booktabs}
\usepackage{array}
\usepackage{adjustbox}
\usepackage[most]{tcolorbox}

\def\BibTeX{{\rm B\kern-.05em{\sc i\kern-.025em b}\kern-.08em
    T\kern-.1667em\lower.7ex\hbox{E}\kern-.125emX}}

\begin{document}

\title{\LARGE \bf
Bridging Policy and Real-World Dynamics: LLM-Augmented Rebalancing for Shared Micromobility Systems
}

\author{Heng Tan$^{1}$, Hua Yan$^{1}$, Yu Yang$^1$
\thanks{$^{1}$Heng Tan, Hua Yan, and Yu Yang are with the Department of Computer Science $\&$ Engineering, Lehigh University, Bethlehem, USA
        {\tt\small het221@lehigh.edu, huy222@lehigh.edu, yuyang@lehigh.edu}}%
\thanks{This work was supported in part by NSF grants 2246080, 2318697, and 2427915. We also thank the anonymous reviewers for their valuable comments and feedback.}
}

\maketitle

\begin{abstract}

Shared micromobility services such as e-scooters and bikes have become an integral part of urban transportation, yet their efficiency critically depends on effective vehicle rebalancing. Existing methods either optimize for average demand patterns or employ robust optimization and reinforcement learning to handle predefined uncertainties. However, these approaches overlook emergent events (e.g., demand surges, vehicle outages, regulatory interventions) or sacrifice performance in normal conditions.
We introduce AMPLIFY, an LLM-augmented policy adaptation framework for shared micromobility rebalancing. The framework combines a baseline rebalancing module with an LLM-based adaptation module that adjusts strategies in real time under emergent scenarios. The adaptation module ingests system context, demand predictions, and baseline strategies, and refines adjustments through self-reflection.
Evaluations on real-world e-scooter data from Chicago show that our approach improves demand satisfaction and system revenue compared to baseline policies, highlighting the potential of LLM-driven adaptation as a flexible solution for managing uncertainty in micromobility systems.

\vspace{6pt}

\end{abstract}

\input{1Introduction}

\input{2RelatedWork}

\input{3Design}

\input{4Evaluation}

\input{5Conclusion}

\bibliographystyle{ieeetr}
\bibliography{refs}

\end{document}

%% file: 1Introduction.tex
\section{Introduction}

Shared micromobility services (e.g., scooters and bikes) have rapidly expanded in urban areas around the world, offering flexible, low-emission, and last-mile transportation options for millions of users~\cite{zhong2023rlife,zhong2024adatrans}. According to the recent reports~\cite{NACTO2023micromobility}, global shared micromobility trips have surpassed billions annually, and cities are increasingly integrating such services into their public transit ecosystems. 
However, a key operational challenge in shared micromobility systems is vehicle rebalancing, i.e., ensuring that vehicles are distributed across space to meet spatiotemporal demand. Without effective rebalancing, certain regions may experience shortages while others face oversupply, resulting in poor user experience and revenue loss for operators~\cite{NACTO2023micromobility}. In this work, we utilie electric scooters (e-scooters) as an example to study the rebalancing problem of shared micromobility vehicles.

In recent years, a variety of rebalancing strategies have been developed for shared micromobility systems. 
These methods can be broadly categorized into uncertainty-ignorant vehicle rebalancing~\cite{tan2025realism,wu2024fleet,he2022socially,tan2025realism,tan2025small} and uncertainty-aware vehicle rebalancing~\cite{li2025robust,zhang2025stochastic,bruck2025robust,he2023robust,he2023robust1,yan2024robust,yang2024mallight}. 
Uncertainty-ignorant methods typically determine rebalancing actions based on the current vehicle distribution and predictions of future demand prior to the next rebalancing interval. To optimize these decisions, researchers have employed techniques such as mixed-integer programming~\cite{guo2020vehicle,yuan2019p} and sequential decision-making frameworks such as reinforcement learning~\cite{he2022socially,tan2023joint,tan2025realism}, with the goal of maximizing system-level performance metrics (e.g., cumulative trip revenue or satisfied user demand). However, these approaches primarily emphasize optimality under general conditions or average demand patterns, while emergent or rare events are often overlooked, such as sudden demand surges~\cite{tan2025small}, unexpected vehicle unavailability~\cite{NACTO2023micromobility}, or newly imposed city-level objectives~\cite{chicago_data}.
To better address such uncertainties, prior studies have designed uncertainty-aware rebalancing methods based on robust optimization~\cite{li2025robust,zhang2025stochastic,bruck2025robust} and learning-based strategies~\cite{he2023robust,he2023robust1}. 

While these methods are effective in handling predefined uncertainties, they face an inherent trade-off between normal and emergent cases: the stronger the robustness to uncertainty, the more conservative the resulting policy becomes~\cite{li2025robust,he2023robust,miao2021data,he2023robust1}, often leading to degraded performance under typical demand conditions.
A practical solution is to introduce human intervention, as explored in related domains such as robotics~\cite{liu2022robot,luo2025precise} and autonomous driving~\cite{wu2021human,huang2024human}, where humans can dynamically adapt policy outputs in emergent situations. However, deployment of such human-in-the-loop approaches in shared micromobility systems remains rare, due to limited human resources, scalability concerns, and the difficulty of integrating human inputs into automated decision-making pipelines.

Recent advances in large language model (LLM)-based agent frameworks have demonstrated their remarkable capabilities in situational reasoning and planning across various domains such as robotics~\cite{zeng2025routine}, gaming~\cite{tan2025llm}, and autonomous systems~\cite{sun2023adaplanner}. 
For instance, LLM agents can be fine-tuned with experience datasets to adapt their actions dynamically to specific task objectives~\cite{yao2023retroformer}. They can also leverage in-context learning and memory-driven reasoning to achieve generalization and scalability without fine-tuning~\cite{shinn2023reflexion}.
Inspired by those works, we explore how an LLM-augmented agent can serve as a decision-making assistant to adapt vehicle rebalancing strategies towards the emergent situations.

We introduce AMPLIFY, an LLM-\underline{a}ug\underline{m}ented po\underline{li}cy adaptation \underline{f}ramework for shared micromobilit\underline{y} vehicle rebalancing. The framework is modular and consists of two decoupled components: (1) a vehicle rebalancing module that generates baseline strategies using either learning-based approaches (e.g., reinforcement learning) or rule-based methods (e.g., optimization); and (2) an LLM-augmented adaptation module that modifies these strategies in response to emergent conditions (e.g., sudden demand surges, event-driven disruptions, or new policy interventions from regulators).
The adaptation module takes as input structured information, including the existing rebalancing strategy (i.e., baseline policy), current vehicle distribution, predicted demand, and a description of emergent conditions.
It then produces an adapted rebalancing plan. 
To improve both reasoning quality and feasibility, we employ a self-reflection mechanism in which the LLM iteratively verifies, revises, and finalizes its initial adaptation. 
By decoupling the baseline policy from the adaptation module, our framework achieves high flexibility and compatibility with diverse rebalancing algorithms. 
It enables real-time, human-like intervention without retraining or encoding scenario-specific rules, offering a generalizable and scalable solution for adapting to uncertainty in shared micromobility systems.

In summary, the key contributions are as follows:
\begin{itemize}
    \item We design a simple yet effective LLM-augmented policy adaptation framework for shared micromobility vehicle rebalancing under emergent scenarios. Unlike traditional approaches that rely on retraining or robust pre-planning, our framework enables real-time strategy adaptation with minimal computational overhead.
    \item We develop a modular system that decouples the vehicle rebalancing backbone (e.g., optimization- or learning-based) from the LLM-augmented agent. The agent reasons over system context and emergent conditions, generates adapted rebalancing strategies, and iteratively refines them through self-reflection.
    \item We conduct comprehensive evaluations using real-world e-scooter data from Chicago. Experimental results show that our framework substantially improves pre-optimized rebalancing strategies in emergent scenarios (e.g., sudden demand surges or vehicle outages), leading to higher system-level performance (e.g., demand satisfaction and revenue). Moreover, we find that grounding the LLM in baseline policies significantly enhances its ability to produce effective and stable adaptations, compared to generating strategies by LLMs directly.
\end{itemize}

%% file: 2RelatedWork.tex
\section{Related Work}
\subsection{Shared Micromobility Vehicle Rebalancing}

A wide range of methods has been developed to optimize vehicle rebalancing strategies in shared micromobility systems, with the goal of maximizing system-level performance such as user satisfaction or operational efficiency. These methods can be broadly categorized into two groups: (i) some researchers assume stable and predictable operational environments and utilize non-uncertainty-aware approaches, including optimization-based methods~\cite{lee2024battery,yun2022automated,yuan2019p} and RL-based methods~\cite{tan2023joint,tan2025realism,zhang2022multi,he2022socially,zhao2024urban}. For example, 
\cite{tan2025realism} proposes a regulatory multi-operator vehicle scheduling framework for shared electric scooters, which incorporates an interaction between a city regulator and multiple shared micromobility operators to balance the tradeoff between city-level goals and operators' interests. However, those methods, which are trained on stable and predictable operational environments, struggle to adapt to these unforeseen scenarios, resulting in suboptimal performance during critical moments. (ii) Other researchers employ uncertainty-aware approaches, which explicitly account for disruptions and variations in demand or supply~\cite{tan2024robust,he2023robust1,li2025robust,zhang2025stochastic}. For example, \cite{li2025robust} develops a distributionally robust optimization approach for vehicle rebalancing in AMoD systems under deep demand uncertainty, explicitly modeling worst-case scenarios to improve resilience. 
However, those methods generally lack the flexibility to adapt to real-time emergent situations, where unexpected demand surges, vehicle failures, or new city-level constraints require immediate and context-aware adaptations to the existing scheduling policy.

\subsection{LLM Agents for Planning}

Recent advances in large language model (LLM)-based agent frameworks have demonstrated their remarkable capabilities in situational reasoning and adaptive planning in complex and dynamic environments~\cite{sun2023adaplanner,zeng2025routine,yao2023retroformer,tan2025llm}. For example, \cite{zeng2025routine} provides a modular planning framework that enhances LLM tool execution accuracy by structuring multi-step workflows. \cite{sun2023adaplanner} introduces closed-loop feedback for dynamic plan refinement, enabling more robust adaptation over long horizons.
However, applying LLMs to transportation systems or vehicle scheduling under emergent situations remains largely underexplored. Our work fills this gap by leveraging an LLM-based agent to adapt pre-optimized vehicle rebalancing strategies in response to emergent situations, demonstrating its potential in real-time operational adaptation.

%% file: 3Design.tex
\section{Design}
\subsection{Problem Description}

\textbf{Problem Setting}: We consider an urban city divided into $N$ regions based on official community boundaries~\cite{chicago_data}, and partition the day into $T$ equal-length time intervals. 
Let $S^{i}_{t}$ denote the number of available shared micromobility vehicles in region $i$ at the beginning of time slot $t$ for $1\leq i\leq N$. 
User demand is defined over origin-destination (OD) pairs. Specifically, $U_t^{i,j}$ denotes the number of trip requests from region $i$ to region $j$ during time slot $t$ for $1\leq i,j\leq N$. Therefore, the overall vehicle distribution and user demand across the city at the beginning of time slot $t$ are denoted respectively as $S_{t} \in \mathbb{N}^N$, and $U_{t} \in \mathbb{N}^{N\times N}$.

\textbf{Rebalancing}: 
To meet future user demand, it is essential for the system operator to redistribute or rebalance shared micromobility vehicles to different areas. We define $a_{t} \in \mathbb{N}^{N \times N}$ $(a_{t}=\{a^{i,j}_{t}\}_{1\leq i,j\leq N})$ as the operator's rebalancing strategy at the beginning of the timeslot $t$. In our work, we modularize the vehicle rebalancing generation, which inputs the current vehicle distribution and predicted user demand of future $h$ time slots. Therefore, the rebalancing strategy $a_{t}$ is formulated as:
\begin{equation}
    a_{t} \leftarrow f_{reb}(S_{t}, U_{t:t+h}),
\end{equation}
where $f_{reb}$ is the vehicle rebalancing policy, which is assumed to be pre-determined by existing works~\cite{tan2023joint,lam2016autonomous}.

\textbf{Adaptation under Emergent Scenarios}:
Although the rebalancing strategy $a_t$ is generated by the pre-determined vehicle rebalancing policy $f_{reb}$, real-world operations often encounter emergent scenarios that violate prior assumptions, such as sudden demand surges, vehicle malfunctions, or regulatory interventions (e.g., constraints).
We define each emergent scenario at time $t$ as a tuple $\mathcal{E}_t = \{ \Delta U_t, \Delta S_t, \mathcal{C}_t \}$, which respectively describes unexpected variations in user demand, vehicle availability, and constraint conditions imposed by the external environment.
To enhance system robustness under such emergent scenarios, an adaptation is needed to update the rebalancing strategy based on real-time observations. Specifically, we define an adaptation policy $\pi_{adj}$ that takes the initial rebalancing strategy $a_t$, the current vehicle state $S_t$, the predicted demand of future $h$ time slots $U_{t:t+h}$, and the emergent situation $\mathcal{E}_t$ as input, and outputs a adapted vehicle rebalancing strategy $\hat{a}_t$:
\vspace{-5pt}
\begin{equation}
    \hat{a}_t \leftarrow \pi_{adj}(a_t, S_t, \mathcal{E}_t).
\end{equation}

\textbf{Objective}:
The goal of adaptation is to ensure that vehicle rebalancing strategies remain effective in the presence of sudden distribution shifts (e.g., demand surges or vehicle unavailability) and evolving system objectives (e.g., a dynamic system-level goal such as maximizing vehicle deployment equity). Therefore, we formalize the adaptation objective as maximizing the overall system utility under emergent situations:
\begin{equation}
    \max_{\pi_{adj}} \; \mathbb{E}_{\mathcal{E}_t \sim \mathcal{P}_{\mathcal{E}}} \left[ \mathcal{R}(\hat{a}_t, \mathcal{E}_t) \right],
\end{equation}
where $\mathcal{P}_{\mathcal{E}}$ is the probability distribution of emergent situations. $\mathcal{R}$ quantifies system-level metrics such as demand satisfaction, system revenue, and vehicle deployment equity under adapted strategies.

\subsection{Design Overview}

\begin{figure}[h]
    \centering
    \includegraphics[width=\linewidth, keepaspectratio=true]{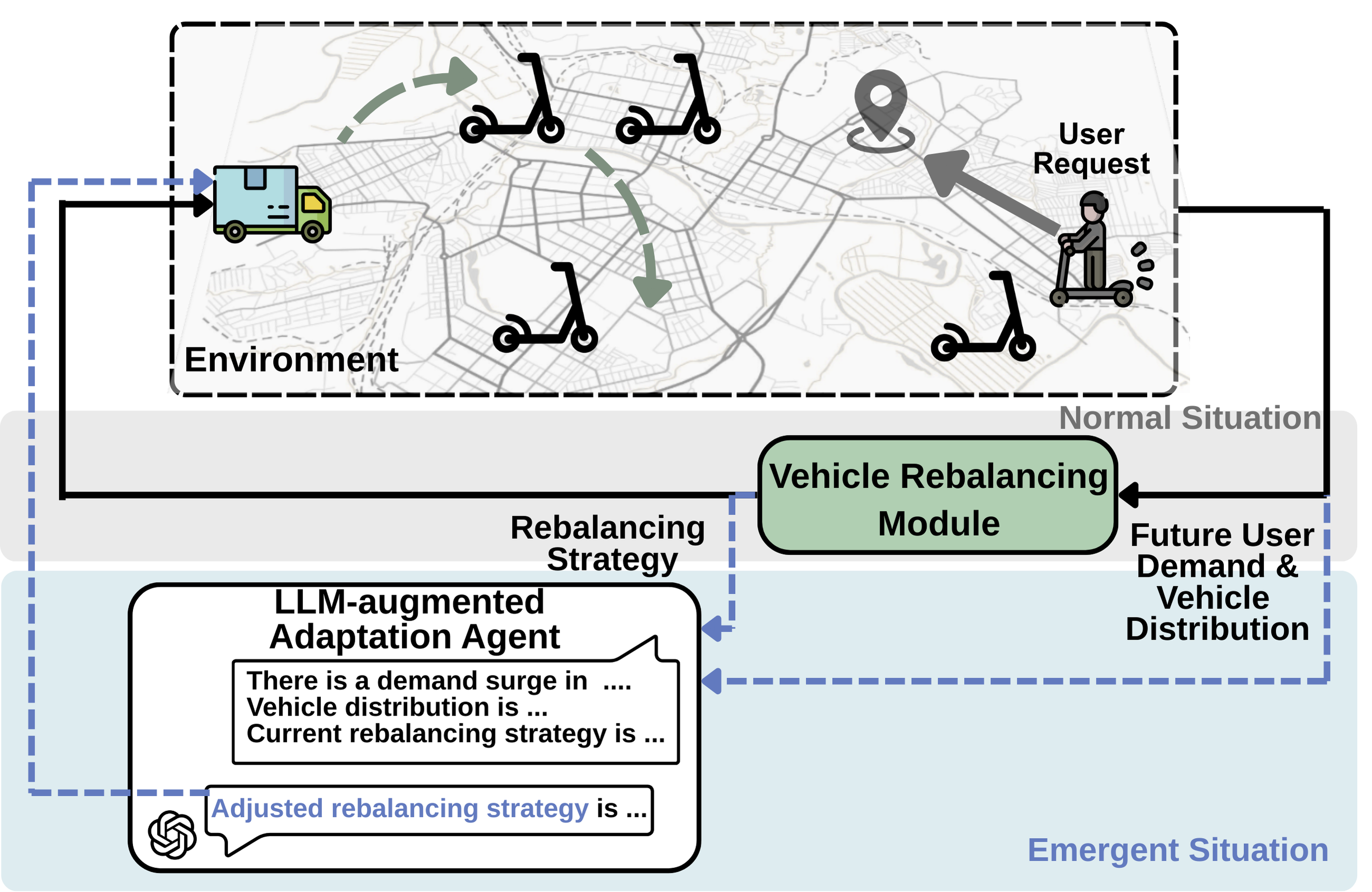}
    \vspace{-10pt}
    \caption{\small LLM-augmented policy adaptation framework for shared micromobility vehicle rebalancing under emergent scenarios}
    \label{fig:framework}
    \vspace{-20pt}
\end{figure}

We design a modular framework for adaptive shared micromobility vehicle rebalancing, which integrates pre-determined vehicle rebalancing strategies with an LLM-augmented agent. As illustrated in Fig.~\ref{fig:framework}, the system consists of three key components:
(1) \textbf{Environment}: It simulates the shared micromobility system operations, including vehicle dynamics, user demand, and request fulfillment.
(2) \textbf{Vehicle Rebalancing Module}: It generates initial rebalancing strategies based on current vehicle distribution and predicted future user demand across all regions. This module can adopt either optimization-based or learning-based algorithms and operates independently of downstream adaptation logic.
(3) \textbf{LLM-augmented Adaptation Agent}: When an emergent situation occurs, it is described in natural language and passed to the LLM agent, together with the current strategy, system-level status (i.e., vehicle distribution and predicted future user demand), and goals (e.g., system revenue). The LLM processes this information and outputs an adapted vehicle rebalancing strategy for execution. To enable more stable and feasible adaptations, there is a self-reflection mechanism that prompts the LLM-augmented agent to verify, revise, and finalize its initial adaptation iteratively.

\subsection{Natural Language Description of Emergent Situations}
In real-world shared micromobility systems, emergent situations—such as sudden demand surges, vehicle unavailability, or dynamic goals—are typically communicated to operators via unstructured textual reports~\cite{ricketts2023scoping,zhang2025developing}. These include messages like ``a large crowd is forming near the stadium", ``vehicles in Region 2 are out of service due to maintenance", or ``ensure better service equity across southern districts". Unlike structured telemetry, these natural language inputs capture real-time, contextualized, and often ambiguous human insights.
To reflect this reality, we adopt \textit{natural language} as the input modality for emergent situations. This enables our LLM-augmented agent to leverage its strong capabilities in understanding and reasoning over natural language inputs, allowing for flexible interpretation and transformation into actionable internal representations. We categorize emergent situations into three major types: 
(1) \textbf{Rising Demand: Sudden Surges in User Requests}. Demand surges represent one of the most frequent and critical emergent situations in shared micromobility systems~\cite{NABSA2024Report,NACTO2023micromobility}. In real-world scenarios, however, such surges are not always directly measurable. To comprehensively assess the robustness of our LLM-augmented agent, we consider two subtypes: (i) \textit{observable surges}, where the magnitude of the increased demand are explicitly available (e.g., from sensors or early warnings), and (ii) \textit{latent surges}, where only vague signals are conveyed via natural language (e.g., ``a large crowd is forming near the stadium'') without precise quantification. These two cases are used to evaluate how well the LLM-augmented agent can infer and adapt to demand shifts under varying degrees of information completeness.
(2) \textbf{Shrinking Supply: Vehicle Shortages and Failures}. Vehicle unavailability frequently arises due to maintenance, vandalism, or battery depletion, especially in shared micromobility systems with large, distributed fleets~\cite{ITF2024Safer,NACTO2023micromobility}. These disruptions are often reported by region staff or monitoring systems in natural language, such as ``bikes in Region 3 are under maintenance.'' Therefore, we introduce vehicle unavailability scenarios to test whether the LLM agent can recognize unavailable supply, avoid infeasible reallocations, and adapt rebalancing accordingly.\\
(3) \textbf{Dynamic Goals}. In addition to maximizing user demand satisfaction, real-world shared micromobility systems are increasingly expected to fulfill additional system-level goals imposed by regulators~\cite{tan2025realism,chicago_data}, such as equity, coverage, or sustainability. In this work, we use vehicle deployment equity as an example, which is ensuring a fair spatial distribution of the demand-supply ratio across regions. Notably, this natural language framing enables the flexible definition and adaptation of equity goals under various metrics (e.g., entropy~\cite{de2025trade}, variance~\cite{tan2025realism}, minimum coverage~\cite{chicago_data}), and can be extended to other system-level goals in future applications.

\subsection{LLM-augmented Rebalancing Strategy Adaptation}
To enable dynamic adaptation of rebalancing strategies in response to emergent scenarios, we employ an LLM as a zero-shot planner that refines pre-generated vehicle rebalancing decisions. Compared with directly using LLMs for planning, our experiments show that incorporating pre-generated strategies significantly improves planning effectiveness and stability.
The LLM is provided with a structured prompt that includes a comprehensive system background, real-time system status, initial vehicle rebalancing strategy, and explicit instructions as shown in Box~1. The goal is to generate an adapted rebalancing plan that is both operationally feasible and aligned with emergent requirements. The prompt is composed of several key sections:
(1) \textbf{Background}: To guide the LLM’s adaptation reasoning, we formulate a structured background prompt that specifies the system state (vehicle availability $V$, demand prediction $D$), the operator’s current rebalancing plan $R$, and the global objective of maximizing net revenue. This structured grounding provides the LLM with the necessary operational context to evaluate trade-offs and generate effective rebalancing adaptations.
(2) \textbf{Real-time Context Input}: To enable the LLM to make informed and context-aware adaptations, we present a natural language encoding of the current system status and the operator's initial vehicle rebalancing plan. Specifically, we include: (i) region-level vehicle supply, (ii) predicted demand flows across origin-destination pairs, and (iii) region-level trip variance statistics derived from historical data. The initial rebalancing strategy generated by the pre-determined vehicle rebalancing policy is also described in natural language form.
(3) \textbf{Adaptation Instruction}: To guide the LLM in making principled and goal-aware rebalancing decisions, we formulate a multi-step adaptation instruction. This instruction prompts the LLM to think through the structured reasoning steps before proposing any adaptation to the initial rebalancing plan, including region-level rebalancing necessity, goal alignment analysis, and adaptation justification.

To further enhance the robustness, feasibility, and multi-goal alignment of the LLM-generated rebalancing strategy adaptation, we incorporate a self-reflection loop that enables the LLM to critically evaluate and revise its own output, inspired by recent advances in reflective prompting and chain-of-thought self-verification~\cite{shinn2023reflexion,sun2023adaplanner}. The self-reflection loop proceeds as follows: 
(1) The LLM receives a full prompt containing system background, current vehicle distribution, initial rebalancing plan, predicted user demand, emergent directive, and a prior adaptation output.
(2) It is instructed to reassess this output by examining key feasibility and goal-alignment criteria, including dimensional validity, conservation constraint, and task satisfaction.
\begin{tcolorbox}[title=Box 1: Prompt for LLM-augmented Rebalancing,label=box:LLM]
\textbf{Background}: You are an intelligent agent embedded in a shared micromobility system. The city is divided into $N$ regions...
\textcolor{gray}{// problem formulation of shared micromobility scheduling}

\textbf{System Status}:
\begin{itemize}
    \item \textbf{System Time}: 00:00
    \item \textbf{Current Vehicle Distribution $(S_{t} \in \mathbb{N}^N$)}: [12, 5, 8, 3, ..., 7] \textcolor{gray}{// Each entry $S^{i}_{t}$ denotes the number of vehicles in region $i$}
    \item \textbf{Predicted future Demand ($U_{t:t+h} \in \mathbb{N}^{N \times N}$)}: Region 1 $\rightarrow$ 2: 10, Region 3 $\rightarrow$ 5: 6, ... \textcolor{gray}{// Each entry $U^{i,j}_{t}$ denotes the number of predicted trips from region $i$ to region $j$}
    \item \textbf{Region-level trip variance}: For each region, the trip variance is provided based on historical trip data: region 1 = \{avg: 8.2, std: 3.5, min: 2, max: 13\}, ... \textcolor{gray}{// statistical patterns of trip demand in each region}
\end{itemize}
\textbf{Initial Rebalancing Strategy ($a_t$)}: Move 2 vehicles from Region 4 to 1, 3 from Region 6 to 3, ... \textcolor{gray}{// Generated by a pre-determined rebalancing policy}

\textbf{Emergent Situation}: The city regulator requests the vehicle deployment equity across all regions...

\textbf{Instruction}:
Your goal is to adapt the operator's vehicle rebalancing strategy under the current emergent situation, considering the provided real-time vehicle distribution, predicted future demand, and historical region-level trip variance. You must think through the following steps before proposing any adaptation:
\begin{itemize}
    \item 1. Determine whether reallocation is needed and from which regions to meet both the requirements of the emergency and the operator.
    \item 2. Explain how you balance revenue maximization against the requirement of the emergency. Please calculate the goal achievements before and after your adaptation if the goal of the emergency is different from that of the operator.
    \item 3. Justify your adaptation with clear reasoning.
\end{itemize}

\textbf{Constraint}: The total vehicle number in the city after your adaptation must be consistent with that before your adaptation.

\textbf{Output Example}:
An example of output format... 
\end{tcolorbox}
(3) The LLM then produces a refined adaptation strategy, which is injected back into the next round of reflection as the new candidate solution.
This self-refinement process improves adaptation quality without requiring external supervision. Empirically, we find that this iterative self-reflection often corrects common issues such as over-allocation, infeasible transfers, or failure to balance multiple goals. The final output is typically more robust, constraint-consistent, and aligned with evolving system requirements.

%% file: 4Evaluation.tex
\section{Evaluation}

\begin{table*}[t]
\centering
\small
\renewcommand{\arraystretch}{1.2}
\caption{\small Performance comparison (i.e., demand satisfaction rate (\%) and equity) of different vehicle rebalancing methods under different emergent scenarios. AMPLIFY denotes our LLM-augmented adaptation of the RECOMMEND rebalancing strategy.}
\label{tb:Performance}
\begin{adjustbox}{max width=1.0\textwidth}
\begin{tabular}{c|cccc|cccc|c}
\toprule
\multirow{2}{*}{\textbf{Method}} &
\multicolumn{4}{c|}{\textbf{Rising Demand~($\uparrow$)}} &
\multicolumn{4}{c|}{\textbf{Shrinking Supply~($\uparrow$)}} &
\multirow{2}{*}{\textbf{Equity~($\uparrow$)}} \\
\cmidrule(lr){2-5}\cmidrule(lr){6-9}
 & \textbf{20\%} & \textbf{50\%} & \textbf{80\%} & \textbf{100\%}
 & \textbf{5\%} & \textbf{10\%} & \textbf{15\%} & \textbf{20\%} \\
\midrule
\textbf{SDSM}
  & 84.12 ($\pm$0.44) & 73.36 ($\pm$0.45) & 66.61 ($\pm$0.47) & 61.45 ($\pm$0.49)
  & 81.10 ($\pm$0.44) & 74.07 ($\pm$0.47) & 66.68 ($\pm$0.49) & 57.01 ($\pm$0.53) 
  & -164.27 ($\pm$2.41) \\
\textbf{GA}
  & 87.94 ($\pm$0.79) & 81.76 ($\pm$0.81) & 75.29 ($\pm$0.82) & 70.46 ($\pm$0.83)
  & 85.97 ($\pm$0.77) & 83.76 ($\pm$0.80) & 80.86 ($\pm$0.84) & 76.35 ($\pm$0.87) 
  & -111.54 ($\pm$2.59) \\
\textbf{RECOMMEND}
  & 93.82 ($\pm$1.34) & 88.47 ($\pm$1.35) & 81.57 ($\pm$1.37) & 80.68 ($\pm$1.41)
  & 91.61 ($\pm$1.33) & 89.04 ($\pm$1.34) & 84.92 ($\pm$1.38) & 80.04 ($\pm$1.43)
  & -81.34 ($\pm$2.92) \\
\textbf{LLM-planning}
  & 91.69 ($\pm$2.95) & 91.50 ($\pm$2.96) & 91.07 ($\pm$2.97) & 90.86 ($\pm$2.99)
  & 84.56 ($\pm$2.04) & 82.65 ($\pm$2.06) & 81.38 ($\pm$2.11) & 79.85 ($\pm$2.16) 
  & -76.38 ($\pm$5.74) \\
\textbf{AMPLIFY (ours)}
  & \textbf{97.34 ($\pm$1.46)} & \textbf{97.22 ($\pm$1.47)} & \textbf{97.10 ($\pm$1.47)} & \textbf{96.87 ($\pm$1.49)}
  & \textbf{94.93 ($\pm$1.35)} & \textbf{93.79 ($\pm$1.38)} & \textbf{91.40 ($\pm$1.41)} & \textbf{89.38 ($\pm$1.46)} 
  & \textbf{-51.79 ($\pm$1.84)} \\
\midrule
\textbf{Average Improvment}   & 8.89\% & 16.05\% & 23.48\% & 27.69\%
  & 10.63\% & 13.85\% & 16.49\% & 21.92\% 
  & 52.22\% \\
\bottomrule
\end{tabular}
\end{adjustbox}
\end{table*}

\subsection{Experiment Setting}
\textbf{Implementation}: We conduct our experiments on a publicly available real-world shared e-scooter dataset from Chicago~\cite{chicago_data}, comprising over 629,000 e-scooter trips operated by Lime, Spin, and Bird between June and September 2022. The dataset contains information such as trip time, distance, operator ID, departure time, and region, among other attributes. The entire city is partitioned into 77 regions based on the existing community divisions in the dataset. A day is divided into 24 hours, with operators' rebalancing occurring every 12 hours. For users' vehicle selections, only nearby vehicles with adequate energy can meet user demand.

We implement our method and baselines with PyTorch 1.9.1, Python-mip 1.14.2, and gym 0.21.0 in a Python 3.7 environment, and we train them on a server equipped with 32 GB of memory and a GeForce RTX 3080 Ti GPU. For LLMs, we choose GPT-4o as our LLM-augmented agent. The maximum number of iterations in self-reflection is set as 10. Hyperparameters for vehicle rebalancing baselines are fine-tuned based on the range in the original papers.

\textbf{Baselines}: We evaluate the performance of our model AMPLIFY with the following vehicle rebalancing baselines: (1) \textbf{SDSM}. It is a rule-based demand-supply matching method where each operator rebalances the vehicles to each region based on the ratio of the historical demand. 
(2) \textbf{GA}~\cite{lam2016autonomous}. It is an optimization-based method that utilizes a genetic algorithm to find the optimal vehicle scheduling strategies to maximize the demand satisfaction rate.
(3) \textbf{RECOMMEND}~\cite{tan2023joint}. It is a state-of-the-art shared electric micromobility vehicle scheduling algorithm that utilizes a MARL-based method to learn the optimal vehicle scheduling polices to maximize the system service performance.
(4) \textbf{LLM-planning}. It is an LLM-augmented agent that directly makes vehicle rebalancing decisions. We adopt the Box 1 prompt without initial rebalancing strategies.

\textbf{Metrics}: The evaluation metrics are as follows: (1) \textbf{Average satisfaction rate}: The average satisfaction rate represents the average ratio of satisfied demand to the total user demand among all the regions. (2) \textbf{Vehicle usage equity}~\cite{tan2025realism}: the metric measures the difference between demand-supply ratios of individual regions and the whole city.

\begin{figure*}[t]
  \centering
  \begin{minipage}[t]{0.24\textwidth}
    \centering
    \includegraphics[width=\linewidth]{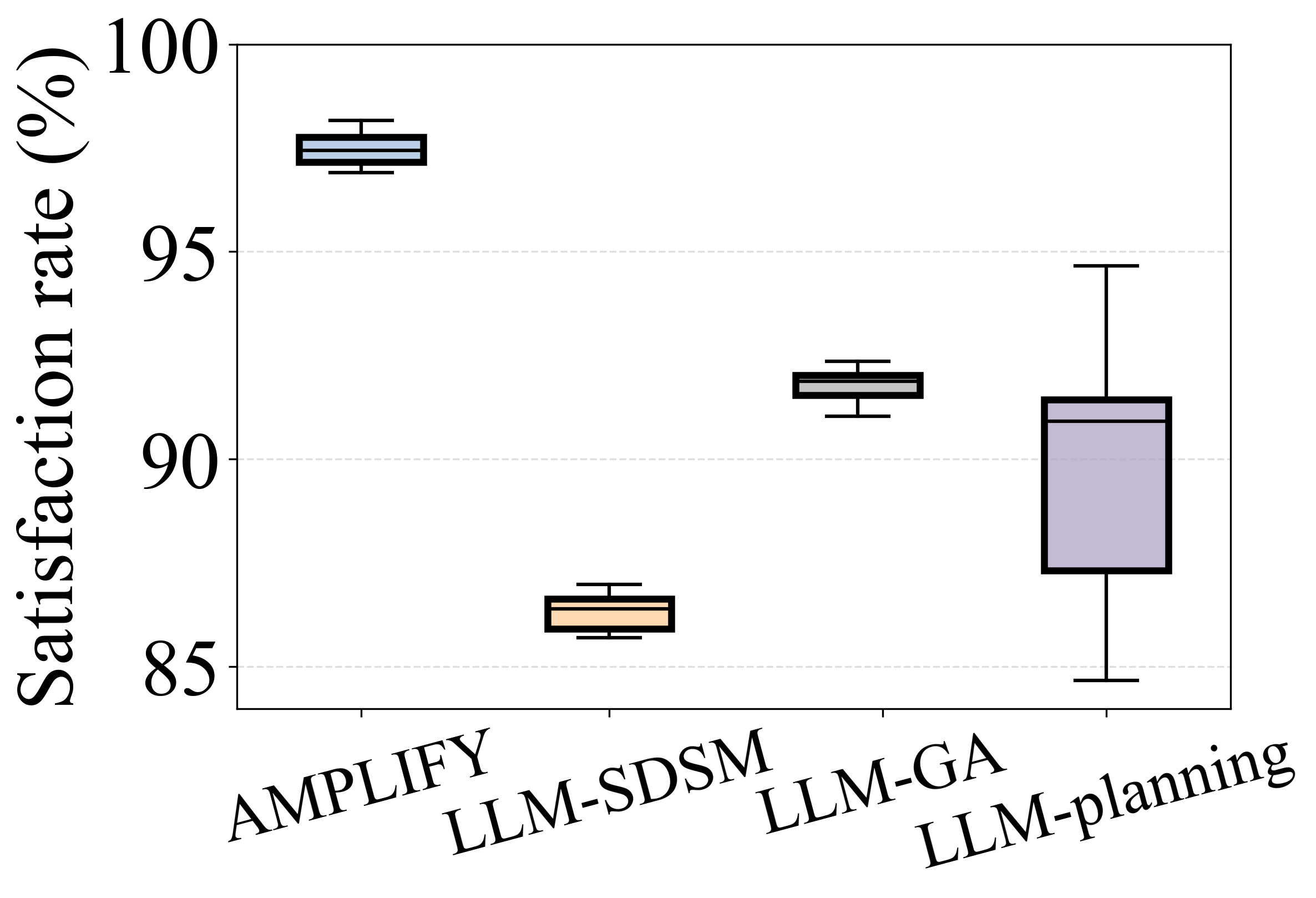}
    \vspace{-10pt}
    
    \caption{Performance under rising demand}
    \label{fig:ab1}
  \end{minipage}
  \hfill
  \begin{minipage}[t]{0.24\textwidth}
    \centering
    \includegraphics[width=\linewidth]{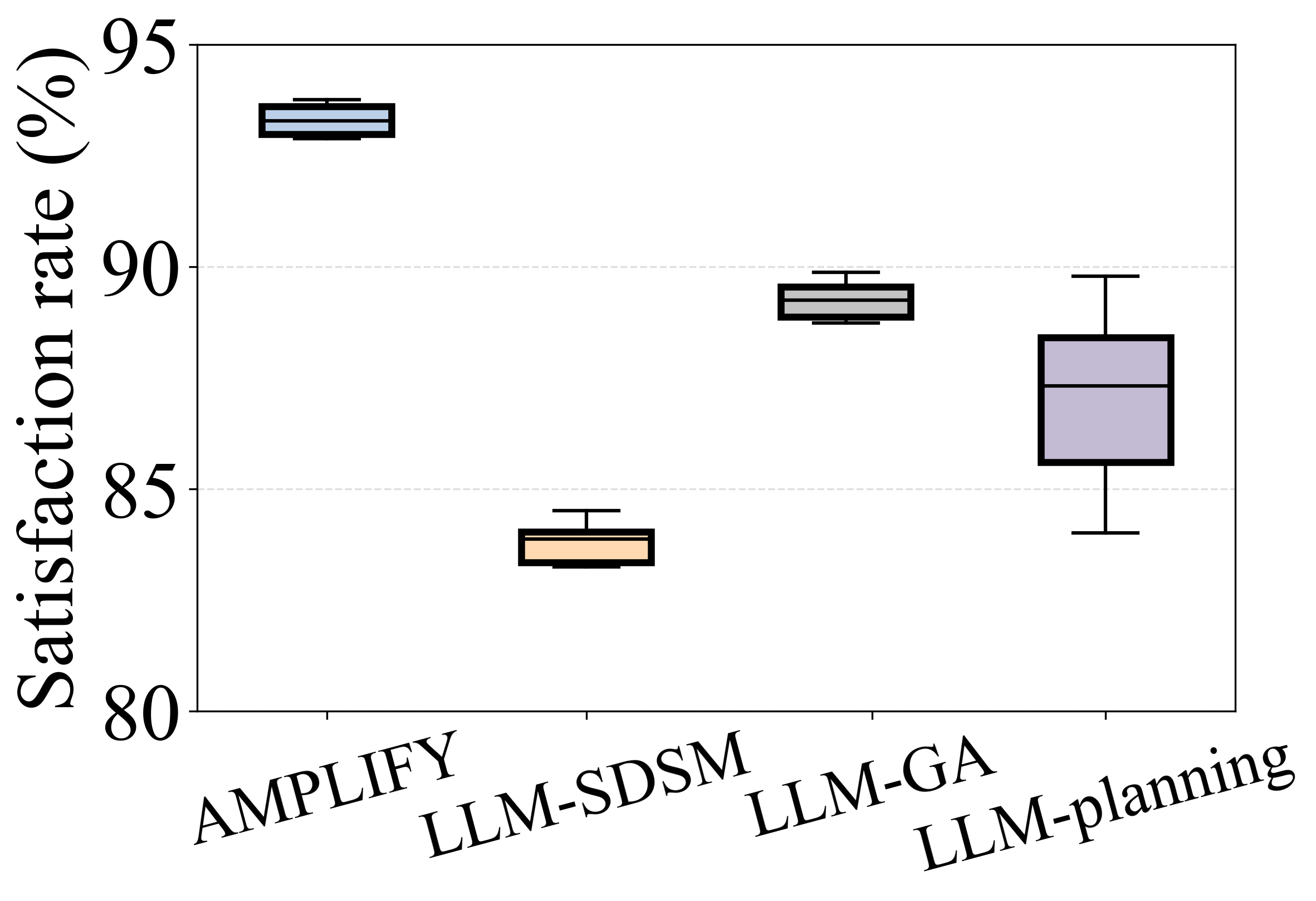}
    \vspace{-10pt}
    
    \caption{Performance under shrinking supply}
    \label{fig:ab2}
  \end{minipage}
  \hfill
  \begin{minipage}[t]{0.24\textwidth}
    \centering
    \includegraphics[width=\linewidth]{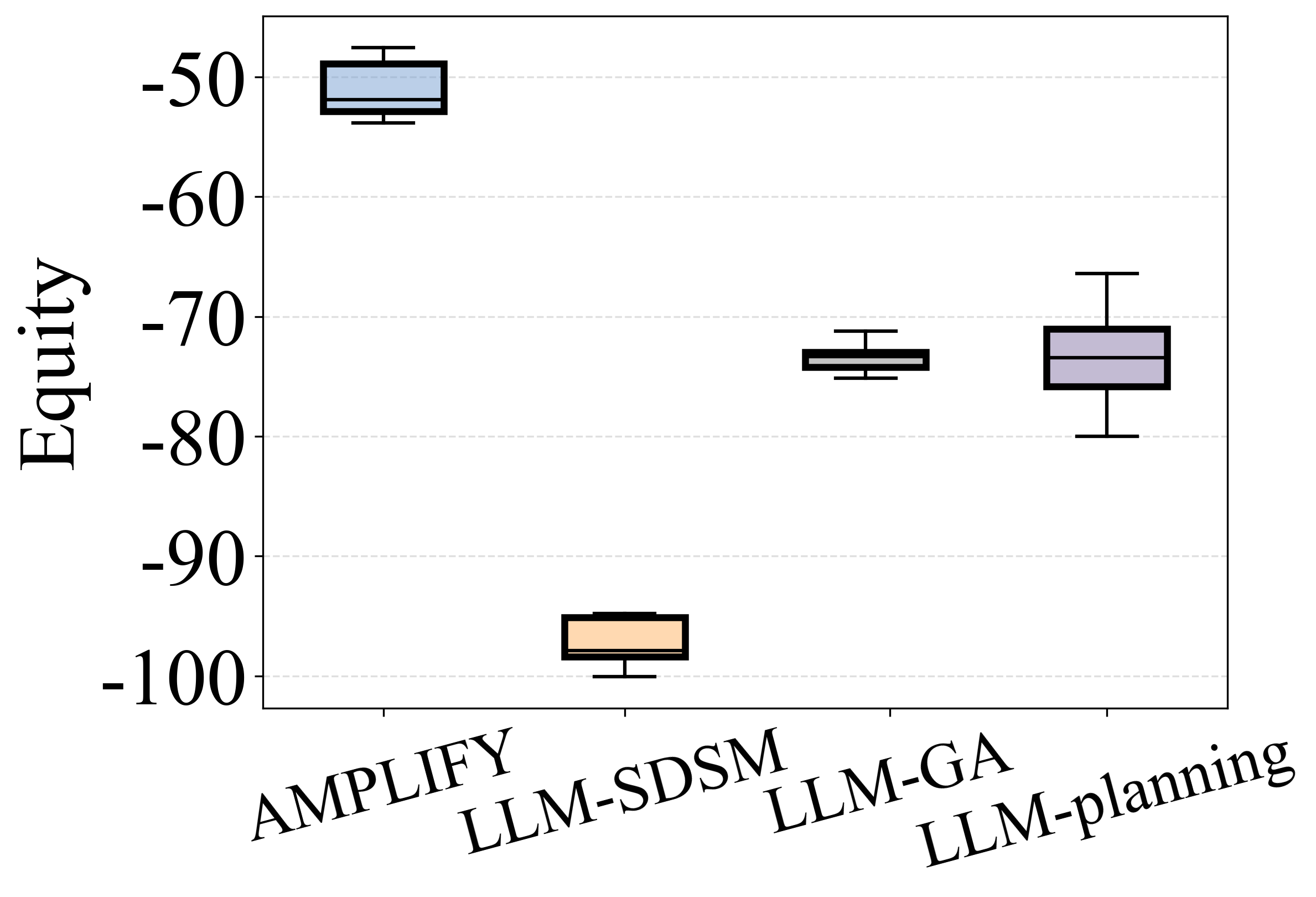}
    \vspace{-10pt}
    
    \caption{Performance under dynamic goals}
    \label{fig:ab3}
  \end{minipage}
  \hfill
    \begin{minipage}[t]{0.24\textwidth}
    \centering
    \includegraphics[width=\linewidth]{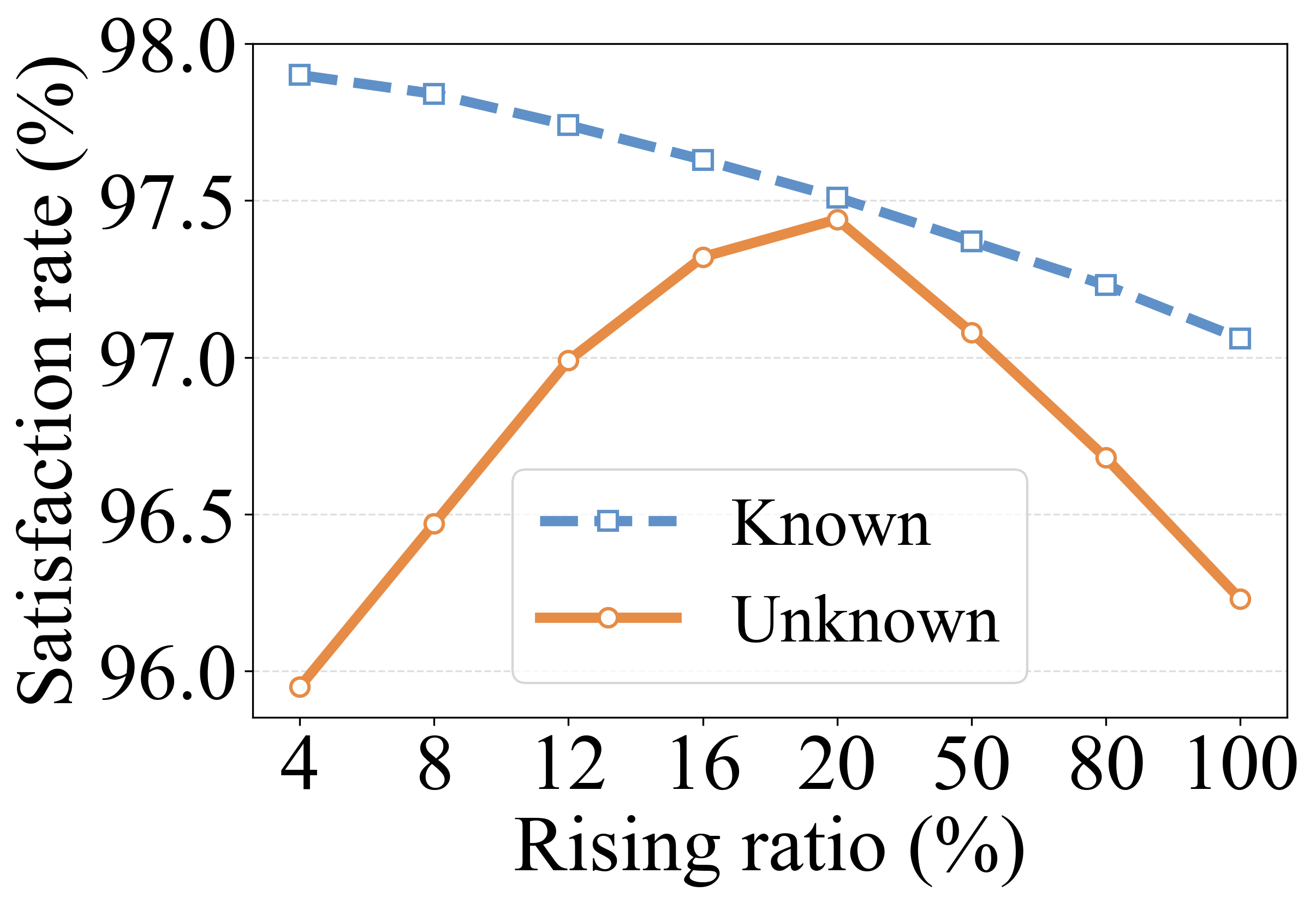}
    \vspace{-10pt}
    \caption{Performance under different scenario descriptions}
    \label{fig:unknown}
  \end{minipage}

  \label{fig:four_in_one_row}
  \vspace{-15pt}
\end{figure*}

\subsection{Overall Performance}
To comprehensively evaluate the adaptability of our LLM-augmented framework, we randomly select ten days in two months as emergent dates and design three distinct types of emergent scenarios that reflect real-world disruptions and policy shifts to evaluate the average performance of different methods under the following different emergent situations: (1) Rising Demand: A sudden surge in user demand across the city, simulated by scaling region-level demand with varying rising ratios ranging from 20\% to 100\%. (2) Shrinking Supply: A reduction in the total available vehicle supply, representing supply-side constraints such as ongoing maintenance. We simulate this by decreasing the available fleet by 5\% to 20\% (randomly select vehicles across all regions). (3) Dynamic Goals: In this setting, the regulator imposes an additional constraint to improve equity in vehicle deployment. We compare our method with other baselines in these three scenarios, and the experiment results (containing average performance with its variance) are shown in Table~\ref{tb:Performance}. In our work, our model is the LLM-augmented adaptation of the RECOMMEND rebalancing strategy, and we discuss the effectiveness of underlying rebalancing strategies in Section~\ref{sec:rs}.

Based on this table, we can know that 
\textbf{(1) Rising Demand:} as user demand increases from 20\% to 100\%, all baseline strategies exhibit a steady decline in satisfaction rate due to increasing service pressure. Notably, AMPLIFY maintains a remarkably stable performance (from 97.51\% to 97.06\%), significantly outperforming all baselines. For example, at 100\% increased demand, AMPLIFY achieves a satisfaction rate of 97.06\%, compared to 90.62\% for LLM-planning and only 63.12\% for SDSM. This result illustrates the effectiveness of our LLM-augmented agent in leveraging initial strategies and promptly redistributing vehicles in response to sudden demand surges. 
\textbf{(2) Shrinking Supply:} Even though AMPLIFY still outperforms all baselines in this scenario, its performance stability reduces compared with that in the scenario of rising demand. The possible reason is the random removal of vehicles across the city, which breaks the original spatial supply distribution and increases the complexity of accurate adaptation for the initial vehicle rebalancing strategy. In contrast, the rising demand scenario benefits from more predictable spatiotemporal patterns, allowing the LLM to generate more effective adaptations even under scaling stress. 
\textbf{(3) Dynamic Goals:} In the equity-focused setting, both the LLM-augmented agent and the LLM-planning baseline (i.e., directly generating rebalancing strategies without any underlying strategy) outperform traditional baselines. This demonstrates that LLMs can not only understand natural language descriptions of complex policy objectives (e.g., minimizing service inequity) but also incorporate these goals into the adaptation of vehicle rebalancing strategies by step-by-step reasoning.

\subsection{Ablation Study}

\subsubsection{The effectiveness of underlying rebalancing strategies}
\label{sec:rs}
To investigate how the choice of underlying rebalancing strategy affects the performance of the LLM-augmented agent, we conduct an ablation study across three representative scenarios: (i) rising demand (+50\%), (ii) shrinking supply (-10\%), and (iii) a dynamic goal (i.e., maximizing vehicle deployment equity). In each scenario, we compare our method with the other three baselines: (i) LLM-planning, (ii) LLM-GA: the LLM adapts vehicle rebalancing strategies generated by a classical genetic algorithm~\cite{lam2016autonomous}, and (iii) LLM-SDSM: the LLM adapts vehicle rebalancing strategies generated by a rule-based method (i.e., SDSM). Each method in each scenario is evaluated by prompting the LLM multiple times to capture the variability in its outputs. We visualize the performance variance of different methods using boxplots, as shown in Fig.~\ref{fig:ab1}, Fig.~\ref {fig:ab2}, and Fig.~\ref {fig:ab3}.

We observe that the effectiveness of the LLM-augmented adaptation agent is highly dependent on the quality of the initial vehicle rebalancing strategy. When provided with a strong baseline (e.g., the RL-based method RECOMMEND), the LLM-augmented adaptation performs exceptionally well, adapting the strategy with high precision and stability. In contrast, when the initial strategy is weak (e.g., the rule-based method SDSM), the LLM-augmented agent struggles to overcome fundamental planning deficiencies, leaving substantial room for improvement. Moreover, in the LLM-planning setting, where no initial strategy is provided, we observe substantial performance variance across all scenarios compared with other baselines that are provided with initial vehicle rebalancing strategies. It highlights the difficulty of open-ended planning tasks for LLMs, especially in constrained domains like vehicle allocation, and indirectly demonstrates that the quality of the initial strategy heavily influences the effectiveness of LLM adaptation.

\subsubsection{The effectiveness of different LLMs}
To evaluate how different LLMs perform in adapting pre-determined vehicle rebalancing strategies to various emergent scenarios, we conduct an ablation study comparing four adaptation models using different language backbones: GPT‑3.5‑turbo, LLaMA‑3.1‑8B, LLaMA‑3.1‑70B, and GPT‑4o (used in AMPLIFY) to adapt vehicle rebalancing strategies generated by RECOMMEND, as shown in Table~\ref{tb:llms}.

Based on the experiment results, we can know that (i) The strategy adaptation capability is tied to the model's reasoning and comprehension capacity: Smaller models such as GPT‑3.5 and LLaMA‑3.1‑8B often fail to generate effective strategy adaptations, generating strategies worse than the original baseline RECOMMEND. In contrast, larger models like LLaMA‑3.1‑70B and GPT‑4o successfully adapt the initial strategy, consistently yielding improved outcomes across all scenarios. (ii) While both larger models (i.e., LLaMA‑3.1‑70B and GPT‑4o) exhibit strong adjustment capability, GPT‑4o (AMPLIFY) delivers the highest improvements, particularly in the scenario of dynamic goals (i.e., equity). This suggests that GPT‑4o possesses stronger reasoning and generalization ability when integrating complex system-level goals into its adaptations.
\vspace*{-5pt}
\begin{table}[h]
\small
\centering
\caption{The performance comparison of the adaptation model with different LLMs}
\captionsetup{font={tiny}}
\label{tb:llms}
\resizebox{\linewidth}{!}{%
\begin{tabular}{cccc}
\toprule
Method  & Rising Demand  & Shrinking Supply & Equity   \\ 
\midrule
GPT-3.5-turbo & 85.51\% & 85.56\% & -104.97 \\ 
Llama-3.1-8B & 85.35\% & 88.74\% & -98.12 \\
Llama-3.1-70B & 93.27\% & 90.17\% & -78.14 \\
GPT-4o & 97.37\% & 93.28\% & -50.48 \\
\bottomrule
\end{tabular}%
}
\vspace*{-5pt}
\end{table}

\subsubsection{The effectiveness of self-reflection}
We find that the self-reflection mechanism is critical for ensuring both the correctness and effectiveness of the LLM-augmented rebalancing strategies. In an ablation study where the self-reflection loop is disabled, the LLM frequently outputs invalid or flawed strategies even when the prompt explicitly encodes key operational constraints. These include: (i) malformed output formats that cannot be parsed into action vectors, (ii) mismatches in total vehicle count before and after adaptation, (iii) violations of basic constraints (e.g., negative vehicle allocations), and (iv) disregard for critical input signals such as emergency needs or equity goals.
To quantify the effectiveness of the self-reflection module, we conduct 20 independent runs of the adaptation process under both settings. Without self-reflection, only 3 out of 20 runs (15\%) yielded valid and usable rebalancing strategies. In contrast, with self-reflection enabled, 17 out of 20 runs (85\%) successfully produced valid, constraint-satisfying, and context-aware strategies.
The experiment results demonstrate that self-reflection acts as a crucial internal verification step, guiding the LLM to revise and validate its reasoning before finalizing an adapted vehicle rebalancing decision.
Nevertheless, we acknowledge that in a small portion (15\%) of cases, the self-reflection mechanism still fails to converge to a valid adaptation. To address these cases and ensure robustness in real deployments, we provide two backup plans: (i) Increase the iteration depth of the self-reflection loop; (ii) Fallback to the initial strategy if no valid adaptation is produced.

\subsection{Impact of Factor}

\subsubsection{The impact of explicit information in emergent scenarios}
To examine the importance of informing the LLM about the specific surge magnitude in emergent demand scenarios, we conduct an ablation study by comparing two variants of our adaptation agent: one that is explicitly prompted with the region-wise rising demand ratio (Known), and one that is only informed of the region-wise rising signal without any surge magnitude information (Unknown).

Based on the experimental results in Fig.~\ref{fig:unknown}, we observe that: (i) even without explicit surge magnitude, the LLM can produce meaningful adaptations by leveraging its internal assumptions. This is because the prompt includes region-level demand variability statistics (mean, standard deviation, min, and max), which implicitly guide the LLM to hypothesize a reasonable demand shift. This capability is evident at the 20\% rising point, where the satisfaction rates for known and unknown variants are nearly identical. 
(ii) Across all rising levels, the known variant consistently outperforms the unknown variant. Not only does it achieve higher satisfaction rates, but it also maintains a more stable performance curve with a lower degradation slope as the rising ratio increases. This indicates that providing precise, structured surge information enables the LLM to generate more targeted and effective rebalancing strategies. 
(iii) The performance degradation of the Known variant reflects the structural limitation of finite vehicle supply: regardless of the LLM’s awareness of surge magnitude, there is a natural bottleneck to how much adapted vehicle rebalancing strategies can serve extreme demand increases.

\subsubsection{The impact of different equity definitions}
To investigate the flexibility of our framework under different equity definitions, we compare two widely used fairness metrics: the Gini coefficient~\cite{welch2013measure} and the Theil index~\cite{de2025trade}. Both metrics quantify the inequality in the spatial distribution of service. They differ in sensitivity and formulation. While the Gini coefficient emphasizes rank-based inequality, the Theil index captures entropy-based disparity and is more sensitive to deviations in the tails. 

As shown in Table~\ref{tb:equity}, prompting the LLM with different equity definitions enables our adaptation agent to generate rebalancing strategies that consistently reduce regional inequality compared with the initial strategies. This demonstrates that the LLM can interpret and adapt to multiple equity formulations expressed in natural language, producing more equitable rebalancing outcomes without retraining. These results highlight the flexibility and effectiveness of our model in aligning outputs with diverse equity objectives.
\vspace{-5pt}
\begin{table}[h] \small \centering
\setlength\tabcolsep{10pt}
\caption{The impact of equity definition}
\captionsetup{font={tiny}}
\vspace{-5pt}
\label{tb:equity}

\begin{tabular}{ccc}
\toprule
Method  & Gini coefficient~($\downarrow$)  & Theil index~($\downarrow$)    \\ 
\midrule
RECOMMEND & 0.6738~($\pm$0.001) & 3.9702~($\pm$0.001)
   \\ 
AMPLIFY     & 0.5007~($\pm$0.022)  & 3.9213~($\pm$0.002) 
 \\ \bottomrule
\end{tabular}

\vspace*{-10pt}
\end{table}

\subsection{Case Study}
\begin{figure}[h]\centering

\begin{minipage}[t]{0.45\linewidth}
    \includegraphics[width=\linewidth, keepaspectratio=true]{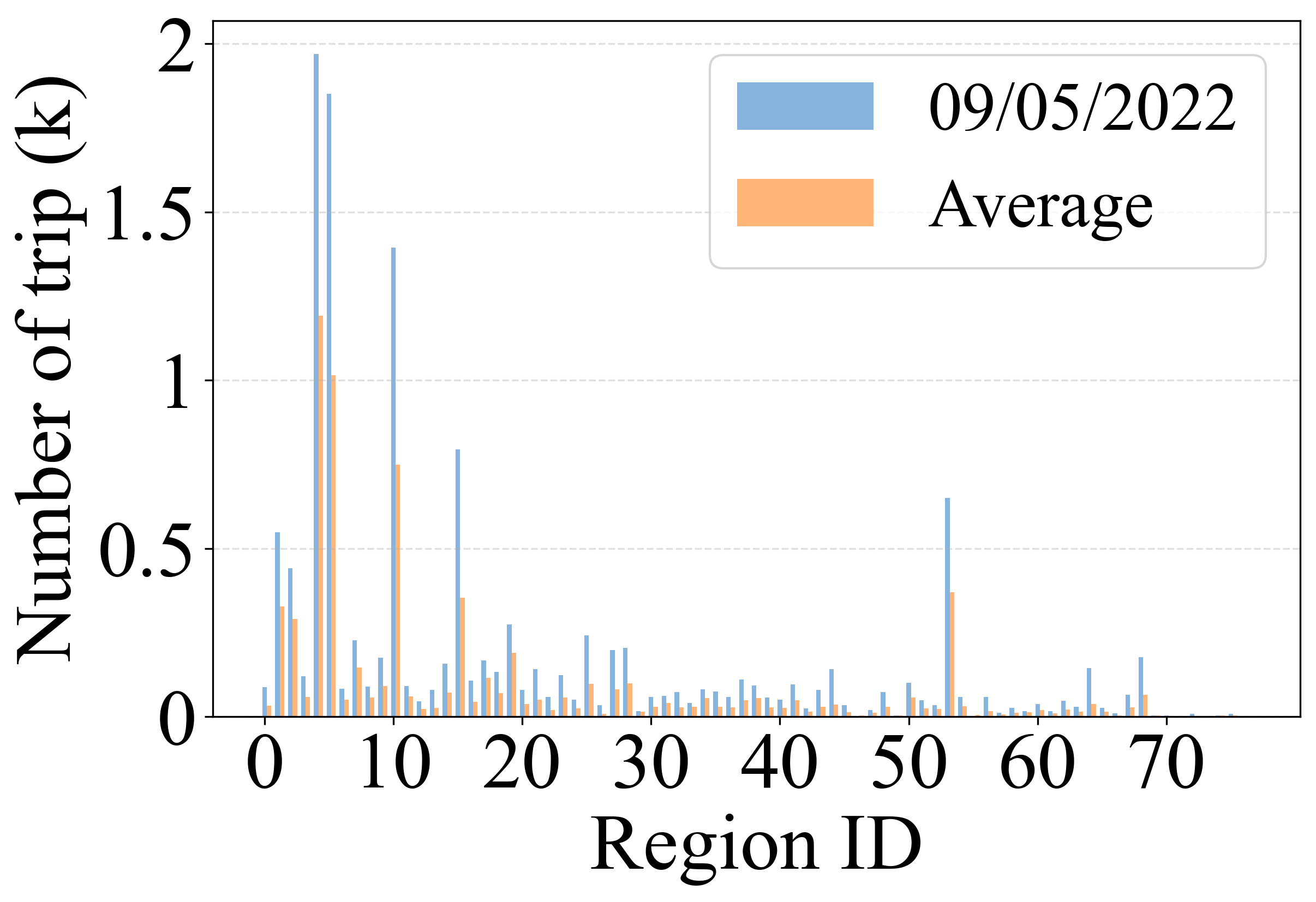}
    \vspace{-15pt}
    \captionsetup{font={small}}
    \caption{The comparison of two demand distributions}
    \label{fig:case2}
\end{minipage}
\hspace{10pt}
\begin{minipage}[t]{0.45\linewidth}
    \includegraphics[width=\linewidth, keepaspectratio=true]{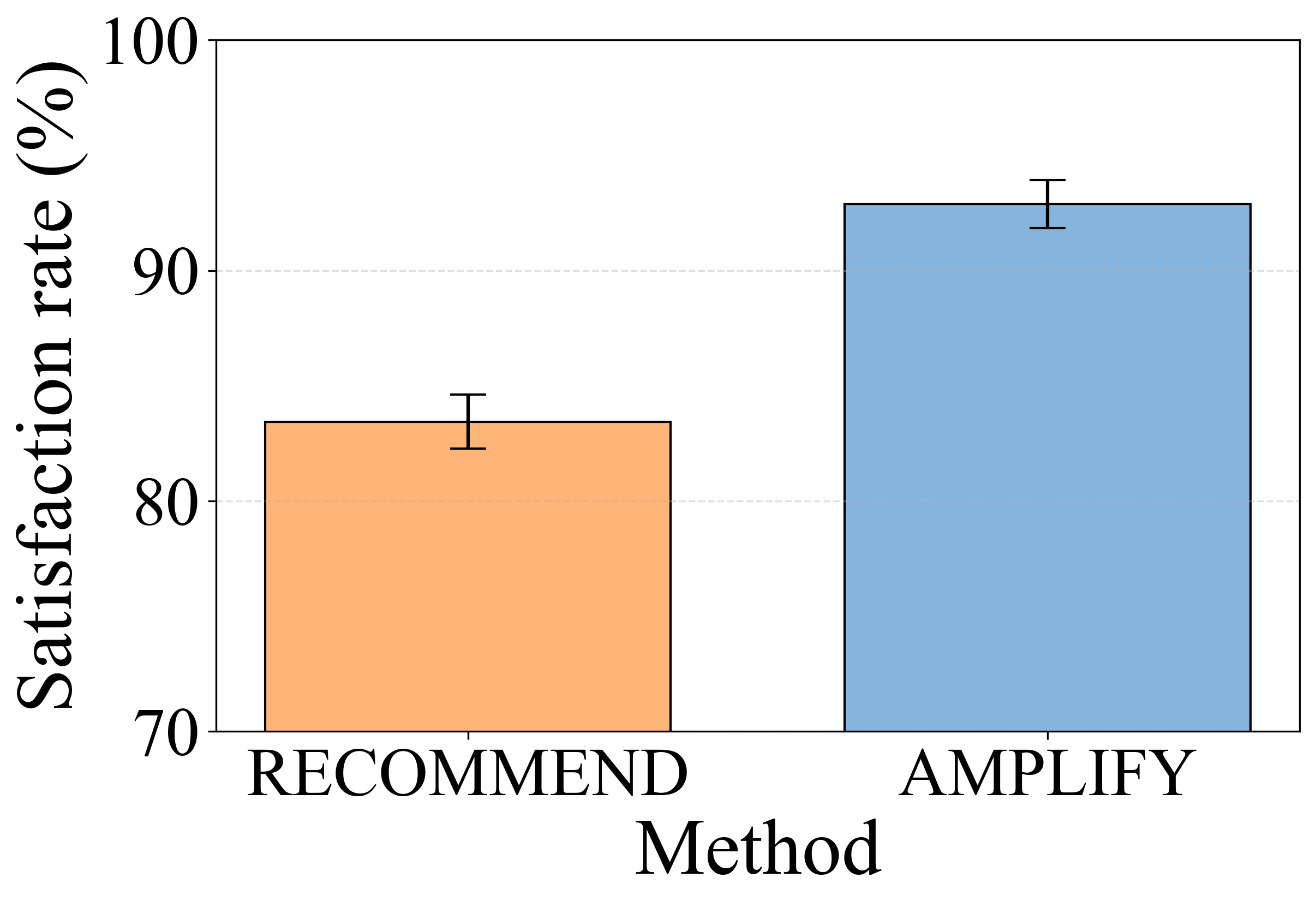}
    \vspace{-15pt}
    \captionsetup{font={small}}
    \caption{The performance comparison of two methods}
    \label{fig:case_study}
\end{minipage}
\vspace{-15pt}
\end{figure}

To evaluate our LLM-augmented vehicle rebalancing strategy under real-world emergent conditions, we select September 5th, 2022 (Monday) in Chicago as a test day. This date coincides with multiple large events, such as Labor Day, the Bitter Jester Music Festival, African Festival of the Arts, and the Taste of Polonia Festival, all contributing to atypical user mobility patterns.
On this day, the system recorded 12,373 shared micromobility trips, which is nearly double the average Monday volume of 6,603 trips. 
Fig.~\ref{fig:case2} presents the demand distribution across all regions for the Monday average and the selected study day, highlighting regions with unusually high demand surges (e.g., region 5 and region 10).

To test whether our model can effectively adapt vehicle rebalancing decisions to such a real-world emergent scenario, we assume that we have prior knowledge of these rising-demand regions before users initiate trips and prompt the context to the LLM-augmented agent to generate the adapted vehicle rebalancing strategy. This is a reasonable assumption since the locations of these events are known. We compare the performance of the adapted vehicle rebalancing strategy with the initial strategy provided by RECOMMEND. The experiment result in Fig.~\ref{fig:case_study} shows that the baseline underperforms on this emergent day, achieving only 83.44\% fulfillment, which is well below its monthly average of 90.61\%. In contrast, our work achieves a significant performance improvement, with a rise to 92.87\%, which demonstrates the potential of our LLM-augmented adaptation framework to adapt rebalancing strategies in real-world emergent settings.

%% file: 5Conclusion.tex
\section{Conclusion}
In this work, we focus on the problem of adapting shared micromobility vehicle rebalancing strategies under various emergent scenarios. We design an LLM-augmented adaptation framework that leverages context-aware prompts and self-reflection to refine vehicle rebalancing strategies provided by pre-determined polices. The evaluation results show that our method consistently improves performance across various settings, achieving at least an 8.89\% improvement in satisfaction rate and 52.22\% in vehicle deployment equity.

%% file: refs.bib
@inproceedings{tan2025realism,
  title={REALISM: A Regulatory Framework for Coordinated Scheduling in Multi-Operator Shared Micromobility Services},
  author={Tan, Heng and Yan, Hua and Yuan, Yukun and Wang, Guang and Yang, Yu},
  booktitle={Proceedings of the 33rd ACM International Conference on Advances in Geographic Information Systems},
  pages={894--904},
  year={2025}
}

@inproceedings{yan2024robust,
  title={Robust route planning under uncertain pickup requests for last-mile delivery},
  author={Yan, Hua and Tan, Heng and Wang, Haotian and Zhang, Desheng and Yang, Yu},
  booktitle={Proceedings of the ACM Web Conference 2024},
  pages={3022--3030},
  year={2024}
}

@inproceedings{zhao2024urban,
  title={Urban-focused multi-task offline reinforcement learning with contrastive data sharing},
  author={Zhao, Xinbo and Zhang, Yingxue and Zhang, Xin and Yang, Yu and Xie, Yiqun and Li, Yanhua and Luo, Jun},
  booktitle={Proceedings of the 30th ACM SIGKDD Conference on Knowledge Discovery and Data Mining},
  pages={4512--4523},
  year={2024}
}

@inproceedings{yang2024mallight,
  title={Mallight: Influence-aware coordinated traffic signal control for traffic signal malfunctions},
  author={Yang, Qinchen and Xie, Zejun and Wei, Hua and Zhang, Desheng and Yang, Yu},
  booktitle={Proceedings of the 33rd ACM International Conference on Information and Knowledge Management},
  pages={2879--2889},
  year={2024}
}

@inproceedings{zhong2023rlife,
  title={Rlife: Remaining lifespan prediction for e-scooters},
  author={Zhong, Shuxin and Yubeaton, William and Lyu, Wenjun and Wang, Guang and Zhang, Desheng and Yang, Yu},
  booktitle={Proceedings of the 32nd ACM International Conference on Information and Knowledge Management},
  pages={3544--3553},
  year={2023}
}

@inproceedings{zhong2024adatrans,
  title={AdaTrans: Adaptive Transfer Time Prediction for Multi-modal Transportation Modes},
  author={Zhong, Shuxin and Wei, Hua and Lyu, Wenjun and Yang, Guang and Hong, Zhiqing and Wang, Guang and Yang, Yu and Zhang, Desheng},
  booktitle={Proceedings of the 33rd ACM International Conference on Information and Knowledge Management},
  pages={3443--3452},
  year={2024}
}

@inproceedings{tan2025small,
  title={Small Fleet, Big Impact: Enhancing Shared Micromobility Efficiency through Minimal Autonomous Vehicle Deployment},
  author={Tan, Heng and Yan, Hua and Yang, Lucas and Yang, Yu},
  booktitle={Proceedings of the 12th ACM International Conference on Systems for Energy-Efficient Buildings, Cities, and Transportation},
  pages={203--212},
  year={2025}
}

@article{yao2023retroformer,
  title={Retroformer: Retrospective large language agents with policy gradient optimization},
  author={Yao, Weiran and Heinecke, Shelby and Niebles, Juan Carlos and Liu, Zhiwei and Feng, Yihao and Xue, Le and Murthy, Rithesh and Chen, Zeyuan and Zhang, Jianguo and Arpit, Devansh and others},
  journal={arXiv preprint arXiv:2308.02151},
  year={2023}
}

@article{shinn2023reflexion,
  title={Reflexion: Language agents with verbal reinforcement learning},
  author={Shinn, Noah and Cassano, Federico and Gopinath, Ashwin and Narasimhan, Karthik and Yao, Shunyu},
  journal={Advances in Neural Information Processing Systems},
  volume={36},
  pages={8634--8652},
  year={2023}
}

@article{li2025robust,
  title={Robust Vehicle Rebalancing with Deep Uncertainty in Autonomous Mobility-on-Demand Systems},
  author={Li, Xinling and Guo, Xiaotong and Wang, Qingyi and Zardini, Gioele and Zhao, Jinhua},
  journal={arXiv preprint arXiv:2507.04520},
  year={2025}
}

@article{tan2025llm,
  title={Llm-guided reinforcement learning: Addressing training bottlenecks through policy modulation},
  author={Tan, Heng and Yan, Hua and Yang, Yu},
  journal={arXiv preprint arXiv:2505.20671},
  year={2025}
}

@misc{NACTO2023micromobility,
  author       = {{National Association of City Transportation Officials (NACTO)}},
  title        = {2023 Shared Micromobility in the {U.S.} and {Canada} Report},
  year         = {2024},
  month        = jul,
}

@article{miao2021data,
  title={Data-driven distributionally robust optimization for vehicle balancing of mobility-on-demand systems},
  author={Miao, Fei and He, Sihong and Pepin, Lynn and Han, Shuo and Hendawi, Abdeltawab and Khalefa, Mohamed E and Stankovic, John A and Pappas, George},
  journal={ACM Transactions on Cyber-Physical Systems},
  volume={5},
  number={2},
  pages={1--27},
  year={2021},
  publisher={ACM New York, NY, USA}
}

@article{yun2022automated,
  title={Automated Mobility-on-Demand Service Improvement Strategy through Latent Class Analysis of Stated Preference Survey},
  author={Yun, Jaewoong and Lee, Jaehyung and Kim, Jinhee},
  journal={Journal of Advanced Transportation},
  volume={2022},
  number={1},
  pages={8281988},
  year={2022},
  publisher={Wiley Online Library}
}

@article{lam2016autonomous,
  title={Autonomous-vehicle public transportation system: Scheduling and admission control},
  author={Lam, Albert YS and Leung, Yiu-Wing and Chu, Xiaowen},
  journal={IEEE Transactions on Intelligent Transportation Systems},
  volume={17},
  number={5},
  pages={1210--1226},
  year={2016},
  publisher={IEEE}
}

@article{lee2024battery,
  title={Battery swapping, vehicle rebalancing, and staff routing for electric scooter sharing systems},
  author={Lee, Gaeun and Lee, Jun Soo and Park, Kun Soo},
  journal={Transportation Research Part E: Logistics and Transportation Review},
  volume={186},
  pages={103540},
  year={2024},
  publisher={Elsevier}
}

@misc{chicago_data,
	author	=	{Chicago.},
        year = {2020},
	title	=	{E-Scooter Trips - 2020 - Chicago Data Portal},
	url	=	{{https://data.cityofchicago.org/Transportation/E-Scooter-Trips-2020/3rse-fbp6/data}}
}

@article{wu2024fleet,
  title={Fleet sizing and static rebalancing strategies for shared E-scooters: A case study in Indianapolis, USA},
  author={Wu, Yuhang and Liu, Tao and Du, Bo},
  journal={Transportation Research Part A: Policy and Practice},
  volume={190},
  pages={104287},
  year={2024},
  publisher={Elsevier}
}

@inproceedings{tan2024robust,
  title={Human Preference-aware Rebalancing and Charging for Shared Electric Micromobility Vehicles},
  author={Tan, Heng and Yuan, Yukun and Yan, Hua and Zhong, Shuxin and Yang, Yu},
  booktitle={2024 IEEE International Conference on Robotics and Automation (ICRA)},
  year={2024},
  organization={IEEE}
}

@inproceedings{he2023robust,
  title={Robust electric vehicle balancing of autonomous mobility-on-demand system: A multi-agent reinforcement learning approach},
  author={He, Sihong and Han, Shuo and Miao, Fei},
  booktitle={2023 IEEE/RSJ International Conference on Intelligent Robots and Systems (IROS)},
  pages={5471--5478},
  year={2023},
  organization={IEEE}
}

@article{zeng2025routine,
  title={Routine: A Structural Planning Framework for LLM Agent System in Enterprise},
  author={Zeng, Guancheng and Chen, Xueyi and Hu, Jiawang and Qi, Shaohua and Mao, Yaxuan and Wang, Zhantao and Nie, Yifan and Li, Shuang and Feng, Qiuyang and Qiu, Pengxu and others},
  journal={arXiv preprint arXiv:2507.14447},
  year={2025}
}

@article{sun2023adaplanner,
  title={Adaplanner: Adaptive planning from feedback with language models},
  author={Sun, Haotian and Zhuang, Yuchen and Kong, Lingkai and Dai, Bo and Zhang, Chao},
  journal={Advances in neural information processing systems},
  volume={36},
  pages={58202--58245},
  year={2023}
}

@article{luo2025precise,
  title={Precise and dexterous robotic manipulation via human-in-the-loop reinforcement learning},
  author={Luo, Jianlan and Xu, Charles and Wu, Jeffrey and Levine, Sergey},
  journal={Science Robotics},
  volume={10},
  number={105},
  pages={eads5033},
  year={2025},
  publisher={American Association for the Advancement of Science}
}

@article{liu2022robot,
  title={Robot learning on the job: Human-in-the-loop autonomy and learning during deployment},
  author={Liu, Huihan and Nasiriany, Soroush and Zhang, Lance and Bao, Zhiyao and Zhu, Yuke},
  journal={The International Journal of Robotics Research},
  pages={02783649241273901},
  year={2022},
  publisher={SAGE Publications Sage UK: London, England}
}

@article{huang2024human,
  title={Human as AI mentor: Enhanced human-in-the-loop reinforcement learning for safe and efficient autonomous driving},
  author={Huang, Zilin and Sheng, Zihao and Ma, Chengyuan and Chen, Sikai},
  journal={Communications in Transportation Research},
  volume={4},
  pages={100127},
  year={2024},
  publisher={Elsevier}
}

@article{wu2021human,
  title={Human-in-the-loop deep reinforcement learning with application to autonomous driving},
  author={Wu, Jingda and Huang, Zhiyu and Huang, Chao and Hu, Zhongxu and Hang, Peng and Xing, Yang and Lv, Chen},
  journal={arXiv preprint arXiv:2104.07246},
  year={2021}
}

@article{bruck2025robust,
  title={The Robust Bike sharing Rebalancing Problem: Formulations and a branch-and-cut algorithm},
  author={Bruck, Bruno P and Coutinho, Walton P and Munari, Pedro},
  journal={European Journal of Operational Research},
  volume={325},
  number={1},
  pages={67--80},
  year={2025},
  publisher={Elsevier}
}

@article{zhang2025stochastic,
  title={Stochastic Optimization under Supply Uncertainty for Multimodal Trip Planning Based on Demand Prediction},
  author={Zhang, Yimeng and Cheng, Jingyi and Cats, Oded and Azadeh, Shadi Sharif},
  year={2025}
}

@inproceedings{he2023robust1,
  title={A robust and constrained multi-agent reinforcement learning electric vehicle rebalancing method in amod systems},
  author={He, Sihong and Wang, Yue and Han, Shuo and Zou, Shaofeng and Miao, Fei},
  booktitle={2023 IEEE/RSJ International Conference on Intelligent Robots and Systems (IROS)},
  pages={5637--5644},
  year={2023},
  organization={IEEE}
}

@inproceedings{tan2023joint,
  title={Joint Rebalancing and Charging for Shared Electric Micromobility Vehicles with Energy-informed Demand},
  author={Tan, Heng and Yuan, Yukun and Zhong, Shuxin and Yang, Yu},
  booktitle={Proceedings of the 32nd ACM International Conference on Information and Knowledge Management},
  pages={2392--2401},
  year={2023}
}

@article{guo2020vehicle,
  title={Vehicle rebalancing with charging scheduling in one-way car-sharing systems},
  author={Guo, Ge and Xu, Tao},
  journal={IEEE Transactions on Intelligent Transportation Systems},
  year={2020},
  publisher={IEEE}
}

@article{welch2013measure,
  title={A measure of equity for public transit accessibility},
  author={Welch, Timothy F and Mishra, Sabyasachee},
  journal={Transportation Research Part A: Policy and Practice},
  volume={88},
  pages={1--13},
  year={2013},
  publisher={Elsevier}
}

@article{de2025trade,
  title={The trade-off between equity and quality in public transportation: lessons from a Brazilian case study},
  author={de Souza Coelho, Pedro and de Albuquerque Nobrega, Rodrigo Affonso and de Oliveira, Leise Kelli and Humberto, Mateus},
  journal={npj Sustainable Mobility and Transport},
  volume={2},
  number={1},
  pages={22},
  year={2025},
  publisher={Nature Publishing Group UK London}
}

@misc{NABSA2024Report,
  author    = {North American Bikeshare \& Scootershare Association (NABSA)},
  title     = {2024 Shared Micromobility State of the Industry Report},
  howpublished = {https://nabsa.net/about/industry/},
  year      = {2024}
}

@techreport{ITF2024Safer,
  author    = {International Transport Forum (ITF), OECD},
  title     = {Safer Micromobility: Technical Background Report},
  institution = {ITF, OECD},
  year      = {2024},
  howpublished = {https://www.itf-oecd.org/sites/default/files/safer-micrombility-technical-report.pdf}
}

@article{zhang2025developing,
  title={Developing real-time IoT-based public safety alert and emergency response systems},
  author={Zhang, Han and Zhang, Runze and Sun, Jiamanzhen},
  journal={Scientific Reports},
  volume={15},
  number={1},
  pages={29056},
  year={2025},
  publisher={Nature Publishing Group UK London}
}

@article{ricketts2023scoping,
  title={A scoping literature review of natural language processing application to safety occurrence reports},
  author={Ricketts, Jon and Barry, David and Guo, Weisi and Pelham, Jonathan},
  journal={Safety},
  volume={9},
  number={2},
  pages={22},
  year={2023},
  publisher={MDPI}
}

@inproceedings{he2022socially,
  title={Socially-Equitable Interactive Graph Information Fusion-based Prediction for Urban Dockless E-Scooter Sharing},
  author={He, Suining and Shin, Kang G},
  booktitle={Proceedings of the ACM Web Conference 2022},
  pages={3269--3279},
  year={2022}
}

@inproceedings{yuan2019p,
  title={p\^{} 2Charging: proactive partial charging for electric taxi systems},
  author={Yuan, Yukun and Zhang, Desheng and Miao, Fei and Chen, Jimin and He, Tian and Lin, Shan},
  booktitle={2019 IEEE 39th International Conference on Distributed Computing Systems (ICDCS)},
  pages={688--699},
  year={2019},
  organization={IEEE}
}

@inproceedings{zhang2022multi,
  title={Multi-agent graph convolutional reinforcement learning for dynamic electric vehicle charging pricing},
  author={Zhang, Weijia and Liu, Hao and Han, Jindong and Ge, Yong and Xiong, Hui},
  booktitle={Proceedings of the 28th ACM SIGKDD conference on knowledge discovery and data mining},
  pages={2471--2481},
  year={2022}
}
